# HYBRIDNET FOR DEPTH ESTIMATION AND SEMANTIC SEGMENTATION

*Dalila Sánchez-Escobedo, Xiao Lin* \*, *Josep R. Casas, Montse Pardàs*

Universitat Politècnica de Catalunya. BARCELONA**TECH**
Image and Video Processing Group
C. Jordi Girona 31, 08034 Barcelona, Spain.

## ABSTRACT

Semantic segmentation and depth estimation are two important tasks in the area of image processing. Traditionally, these two tasks are addressed in an independent manner. However, for those applications where geometric and semantic information is required, such as robotics or autonomous navigation, depth or semantic segmentation alone are not sufficient. In this paper, depth estimation and semantic segmentation are addressed together from a single input image through a hybrid convolutional network. Different from the state of the art methods where features are extracted by a sole feature extraction network for both tasks, the proposed HybridNet improves the features extraction by separating the relevant features for one task from those which are relevant for both. Experimental results demonstrate that HybridNet results are comparable with the state of the art methods, as well as the single task methods that HybridNet is based on.

## 1. INTRODUCTION

Semantic segmentation and depth information are intrinsically related and both pieces of information need to be considered in an integrated manner to succeed in challenging applications, such as robotics [1] or autonomous navigation [2]. In robotics, performing tasks in interactive environments requires to identify objects as well as their distance from the camera. Likewise, autonomous navigation applications need a 3D reconstruction of the scene as well as semantic information, to ensure that the agent device has enough information available to carry out the navigation in a safe and independent manner. Although RGB-D sensors are currently being used in many applications, most systems only provide RGB information. This is why addressing depth estimation and semantic segmentation under a unified framework is of special interest.

In the last years, deep learning techniques have shown extraordinary success for both tasks. This paper introduces a hybrid convolutional network that integrates depth estimation and semantic segmentation into a unified framework. The idea of integrating the two tasks into a sole structure is motivated by the fact that segmentation information and depth maps represent geometrical information of a scene. In this paper we propose to build a model where the features extracted are suitable for both tasks, thus leading to an improved accuracy in the estimated information. One of the main advantages of the proposed approach is the straightforward manner semantic segmentation and depth map are estimated from a single image, providing a feasible solution to these problems.

\*corresponding author

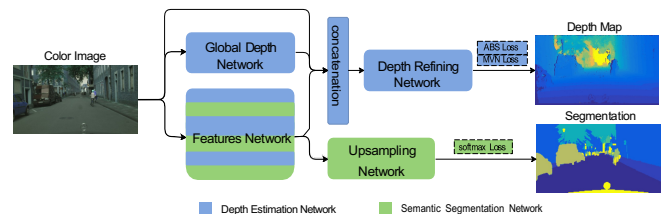

**Fig. 1**: **HybridNet.** Overview of the proposed hybrid convolutional framework, consisting of two main parts: "Depth estimation network"(blue) and "Semantic segmentation network" (green). Notice both networks are linked in the features network block (blue&green).

## 2. RELATED WORK

Deep learning techniques address depth estimation and semantic segmentation problems with efficient and accurate results. One of the most influential approaches on semantic segmentation is the DeepLab model presented in [3]. This model integrates a VGG structure [4] for the features extraction and upsamples the feature maps with atrous convolution layers to obtain the pixel-level semantic labelling. The VGG structure proposed to increase the receptive field of the CNN by stacking as many convolutional layers as needed but keeping the size of the filters $3 \times 3$. It is proved that a significant improvement can be achieved with this deeper network.

Ghiasi et al. [5] present a Laplacian pyramid for semantic segmentation refinement incorporating, into the decoding step, the spatial information contained in the high-resolution feature maps to keep the spatial information destroyed after pooling. Thus, a better dense pixel-accurate labeling is obtained. On the other hand, [6] introduces "DeepLabv3" model which solved the spatial accuracy problem using atrous convolution to capture multi-scale context by adopting multiple atrous rates. Although this approach captured contextual information effectively and fit objects at multiple scales, it still delivers feature maps with low resolution which generate unclear boundaries.

For the depth estimation task from monocular images, one of the first efforts was made by Eigen et al in [7]. This approach estimates a low resolution depth map from an input image as a first step, then finer details are incorporated by a fine-scale network that locally refines the low resolution depth map using the input image as a reference. Additionally, the authors introduced a scale-invariant error function that measures depth relations instead of scale. Ivanecký [8] presents an approach inspired in [7], incorporating estimated gradient information to improve the fine tuning stage. Additionally, in this work a normalized loss function is applied leading to an improvement in depth estimation.

On the other hand, there are some approaches that have addressed depth estimation and semantic segmentation into multiple tasks frameworks. In [9] a unified framework was proposed that incorporates global and local prediction under an architecture that learns the consistency between depth and semantic segmentation through a joint training process. Another unified framework is presented in [10] where depth map, surface normals and semantic labeling are estimated. The results obtained by [10] outperformed the ones presented in [7] proving how the integration of multiple tasks into a common framework may lead to a better performance of the tasks.

A more recent multi-task approach is introduced in [11]. The methodology proposed in this work makes initial estimations for depth and semantic label at a pixel level through a joint network. Later, depth estimation is used to solve possible confusions between similar semantic categories and thus to obtain the final semantic segmentation. Another multi-task approach by Teichmann et al. [12] presents a network architecture named MultiNet that can perform classification, semantic segmentation and detection simultaneously. They incorporate these three tasks into a unified encoder-decoder network where the encoder stage is shared among all tasks and specific decoders for each task producing outputs in real-time. This work efforts were focused on improving the computational efficiency for real-time applications as autonomous driving. A similar approach is Pixel Level enconding and Depth Layering (PLEDL) [13], this work extended a well known fully convolutional network [14] with three output channels jointly trained to obtain pixel-level semantic labeling, instance-level segmentation and 3D depth estimation.

Multi-task approaches seek to extract features suitable to perform diverse tasks at a time. However, most of them unify these tasks under a feature extraction block whose output becomes the input of a group of decoders designed to carry out each task. We propose a new approach that, additionally to the common features extraction, uses s a global depth estimation network to estimate separately the global layout of a scene from the input image, as shown in figure 1. The main motivation to incorporate this extra step is based on the idea that the features network will focus better on extracting common features working for both tasks by separating the global information extraction only needed in the depth estimation task during the training process. The modularized features extraction process helps on producing better features, which leads to an improved refined depth map and segmentation. The experiments presented in this paper demonstrate that incorporating this additional step improves the results obtained by those where features are extracted by a sole feature extraction network for both tasks [13].

## 3. HYBRIDNET

Our HybridNet model, presented in figure 1, is divided in two main components: The Depth Estimation Network (blue) and The Semantic Segmentation Network (green). The first one initially estimates a depth map of the scene at a global level from the input image via a global depth network. Meanwhile, from the input image, robust feature information is extracted by the features network, which is the component shared between the depth network and the semantic segmentation network. The features network in our hybrid model is based on VGG-net [4]. Finally, with the estimated feature information and the input image, the global depth map is locally refined in the refining depth network obtaining the final depth map. The Depth Estimation Network is based on DepthNet [8]. However, DepthNet is formed by global depth estimation, gradient estimation and depth refinement networks trained separately, whilst Depth Estimation Network is trained end to end skipping gradient estimation.

On the other hand, in the Semantic Segmentation Network robust feature information is obtained by the features network from the input image, as in the depth estimation network. Afterwards, the Upsampling Network estimates a class score map where the number of channels is equal to the number of labels. The upsampling network is based on Atrous Spatial Pyramid Pooling (ASPP) proposed in [3]. We denote the semantic segmentation network as DeepLab-ASPP following [3].

In our proposed system the multi-task work is concentrated in the features network, which is common to both tasks, and is based, in our current implementation, in VGG-net. During the training process, the parameters in the features network are learned in such a way that the extracted features convey both depth and semantic information. As these two pieces of information are complementary for scene understanding, our hybrid model, besides solving the two tasks at a time, can outperform the independent solving of the two tasks, leading to a mutual benefit in terms of accuracy.

Our approach seeks to analyze the common attributes between tasks as well as their distinctions in order to clarify how these two tasks may help each other. The motivation of building a hybrid architecture where each component has a specific function relies on the idea that each part can be replaceable for stronger architectures with more parameters working for the same purpose, in order to obtain expected improvements when more computing resources are available. For example, replacing VGG-net for RES-Net101 [15] in the features network. It is important to remember that the main goal of our hybrid model is to solve more than one task at a time, but above all to find a way to outperform the results of both tasks when addressed separately. This paper considers that sharing parameters between tasks during the training process may lead to mutual benefit in terms of accuracy. Experimental results illustrate the promising performance of our approach compared with similar state of the art methods.

## 4. TRAINING PARAMETERS AND INITIALIZATION

The database used for training and experimental evaluation of our model is the Cityscapes dataset [16] which contains 5000 RGB images manually selected from 27 different cities. The 5000 images of the dataset are split into 2975 training images, 500 images for test validation and, for benchmarking purposes, 1525 images.

We fist train DepthNet [8] with Cityscapes dataset for initialization during 100K iteration. After that we took the parameters of the global depth network and depth refinement network to initialize those blocks in our hybrid model. Features network and upsampling network are initialized with the model provided by DeepLab [3] which was pre-trained for classification purposes on ImageNet. Once we have a good initialization for each block of our hybrid model the whole network is trained ent-to-end using Cityscapes dataset.

The loss function used in the semantic segmentation network $L_S$ is the sum of the cross-entropy terms for each spatial position in the output class score map, being our targets the ground truth labels. All positions and labels of the output class score map are equally weighted in the overall loss function with the exception of those unlabeled pixels which are ignored. The loss function utilized for the depth estimation network is composed of two Euclidean losses. $L_{DL}$ computes Euclidean distance between the ground truth and the estimated depth map in linear space, while the $L_{DN}$ computes the Euclidean distance between the normalized ground truth and the estimated map, both normalized by the mean variance normalization. The hybrid loss function $L_H$ is defined as $L_H = \alpha L_S + (L_{DL} + L_{DN})$, where $\alpha$ is the term used to balance the loss functions of depth estimation and semantic segmentation tasks. For training our hybrid model we defined $\alpha = 1000$.

|  | G | C | IoUclass |
|---|---|---|---|
| HybridNet | **93.26** | **79.47** | **66.61** |
| PLEDL [13] | - | - | 64.3 |
| DeepLab-ASPP [3] | 90.99 | 74.88 | 58.02 |
| FCN [14] | - | - | 65.3 |
| SegNet[17] | - | - | 57.0 |
| GoogLeNetFCN[18] | - | - | 63.0 |

**Table 1**: Evaluation of HybridNet against Multi-task and single task approaches.

## 5. EXPERIMENTS

In this section we present the evaluation of the proposed hybrid model. We aim to determine if the features obtained in the shared part of the HybridNet solving the two tasks simultaneously provide better results than the ones that we would obtain using two identical networks trained separately. This is why in addition to the results of our HybridNet and for comparison purposes, we present the results obtained by the models that solve these two tasks separately.

The models used to perform semantic segmentation and depth estimation independently are DeepLab-ASPP [3] and Depth Net [8], respectively. We trained these two models using the code provided by the authors and the Cityscapes dataset in order to compare results. Likewise, to match the size of the input image with the size that our hybrid model supports, the images in the evaluation set were manually cropped into 18 different images. Once the semantic segmentation and depth estimation were performed, those 18 images were rearranged into the original image size.

Figure 2 provides 4 examples from the evaluation set for visual comparison between the results obtained by our hybrid model and ground truth as well as those obtained by DeepLab-ASPP. The purpose of this figure is to depict the differences between a single task and a multi-task approach. Figure 2 shows how the segmentation performed by HybridNet retains with a greater detail the geometrical characteristics of the objects contained in the scene. Like, for instance, in the 3rd row where the shapes of a pedestrian and a car can be better distinguished in the estimation obtained by the proposed hybrid model than the obtained by DeepLab-ASPP.

For depth estimation evaluation, in figure 2 we present a visual comparison of the results obtained by our hybrid model as well as those obtained by the single task approach Depth Net presented in [8] against the ground truth. Note how the results obtained by our HybridNet seem more consistent with the ground truth than those obtained by Depth Net in terms of the depth layering.

In addition to qualitative results, table 1 presents in the first section a comparison between HybridNet and the Pixel Level enconding and Depth Layering (PLEDL) approach proposed in [13], while the second section presents a compari-son between HybridNet and single-task approaches [14, 17, 3, 13]. Three commonly measures for segmentation performance are used: the global accuracy (G) that counts the percentage of pixels which are correctly labeled with respect to the ground truth labelling, the class average accuracy (C) that determines the mean of the pixel accuracy in each class and mean intersection over union (IoUclass) that measures the average Jaccard scores over all classes. This table shows that HybridNet outperforms the results obtained by DeepLab-ASPP which is the model HybridNet is based on, as well as presenting competitive results related to the state of the art. The evaluation is performed on validation set and test set of cityscapes.

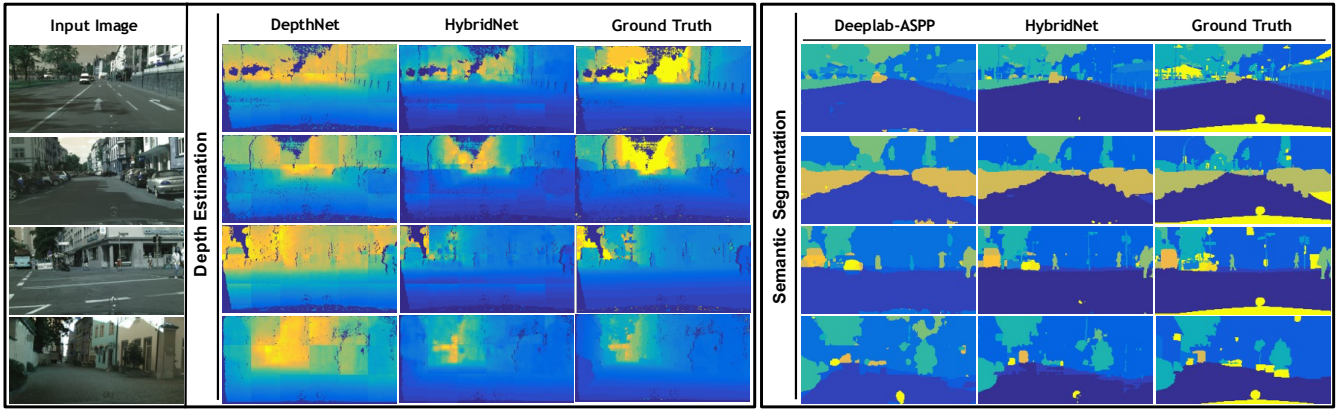

**Fig. 2**: **Qualitative results.** Input image is presented in column 1. Depth estimation results are presented in columns 2 and 3 while column 4 depicts depth ground truth. On the other hand, columns 5 and 6 depict semantic segmentation results while semantic segmentation ground truth is presented in column 7.

|  | HybridNet | DepthNet [8] |  |
| --- | --- | --- | --- |
| $\gamma < 1.25$ | 0.5968 | **0.6048** | higher is better |
| $\gamma < 1.25^2$ | **0.8221** | 0.8187 |  |
| $\gamma < 1.25^3$ | **0.9194** | 0.9152 |  |
| ARD | 0.24 | **0.23** | lower is better |
| SRD | **4.27** | 4.43 |  |
| RMSE-linear | **12.09** | 12.35 |  |
| RMSE-log | 0.4343 | **0.4340** |  |
| SIE | 0.26 | **0.25** |  |

**Table 2**: **Depth estimation.** Quantitative evaluation: ratio threshold, ARD, SRD, RMSE-linear, RMSE-log and SIE.

On the other hand, for depth estimation evaluation we employ 6 commonly used measures: ratio threshold $\gamma$, Absolute Relative Difference (ARD), Square Relative Difference (SRD), Linear Root Mean Square Error (RMSE-linear), Log Root Mean Square error (RMSE-Log) and Scale Invariant Error (SIE) [7]. Table 2 shows the results of our method and those obtained by DepthNet with the different metrics introduced above. Depth Net is identical to the depth estimation part of HybridNet. It is trained with the same training images and configuration used for training the HybridNet. The hybrid architecture outperforms in 4 out of the 8 measures, which proves that training the feature extraction network for the tasks of semantic segmentation and depth estimation simultaneously improves also the depth estimation results. Although the quantitative results presented for depth estimation in PLEDL [13] are similar than ours, they are not comparable since PLEDL only extracts depth for the detected instances, and not the whole scene.

## 6. CONCLUSIONS

This article has introduced a methodology that unifies under a single convolutional framework depth estimation and semantic segmentation tasks using as an input a single image. The main goal of the proposed method is to seek for a better hybrid architecture of convolutional neural networks that modularises the features extraction process by separating it into distinct features extraction for a specific task and common features extraction for both tasks. In this manner, both tasks can benefit from the extracted common features without being affected by those features only relevant to one task, which leads to a better performance. We also prove that solving correlated tasks like semantic segmentation and depth estimation together can improve the performance of methods tackling the tasks separately. The qualitative and quantitative results shown in section 5 illustrate that our hybrid model outperforms the state of the art multi-task approach proposed in [13], as well as the single task approaches it is based on.

## 7. ACKNOWLEDGMENT

This work has been developed in the framework of project TEC2016-75976-R and TEC2013-43935-R, financed by the Spanish Ministerio de Economia y Competitividad and the European Regional Development Fund (ERDF).